\documentclass[letterpaper]{article} 
\usepackage{aaai25}  
\usepackage{times}  
\usepackage{helvet}  
\usepackage{courier}  
\usepackage[hyphens]{url}  
\usepackage{graphicx} 
\usepackage{natbib}  
\usepackage{caption} 
\usepackage{algorithm}
\usepackage{algorithmic}

\usepackage{newfloat}
\usepackage{listings}
\DeclareCaptionStyle{ruled}{labelfont=normalfont,labelsep=colon,strut=off} 
\floatstyle{ruled}
\newfloat{listing}{tb}{lst}{}
\floatname{listing}{Listing}

\newcommand\nnfootnote[1]{%
  \renewcommand\thefootnote{}\footnote{#1}%
  \addtocounter{footnote}{-1}%
}
\renewcommand{\thefootnote}{}

\usepackage{amsmath}
\usepackage{cleveref} 
\usepackage{afterpage}
\usepackage{amssymb}
\usepackage{multirow}
\usepackage{booktabs}
\usepackage{pifont}
\usepackage{colortbl}
\usepackage[table,xcdraw]{xcolor}

\newcommand{\ie}{\textit{i}.\textit{e}.}

\crefname{figure}{Fig.}{Fig.}
\crefname{table}{Tab.}{Tab.}
\crefname{equation}{Eq.}{Eq.}
\crefname{section}{Sec.}{Sec.}

\title{Rethinking High-speed Image Reconstruction Framework with Spike Camera}
\author{Kang Chen\textsuperscript{\rm 1,2}, 
    Yajing Zheng\textsuperscript{\rm 1,2}\footnotemark[1],
    Tiejun Huang\textsuperscript{\rm 1,2,3},
    Zhaofei Yu\textsuperscript{\rm 1,3}\footnotemark[1]
    }
    \affiliations{
    \textsuperscript{\rm 1}School of Computer Science, Peking University\\
    \textsuperscript{\rm 2}State Key Laboratory for Multimedia Information Processing, Peking University\\
    \textsuperscript{\rm 3}Institute for Artificial Intelligence, Peking University
    \\
{mrchenkang@stu.pku.edu.cn, \{yj.zheng,tjhuang,yuzf12\}@pku.edu.cn}
}

\usepackage{bibentry}

\begin{document}

\maketitle
\begin{abstract}
Spike cameras, as innovative neuromorphic devices, generate continuous spike streams to capture high-speed scenes with lower bandwidth and higher dynamic range than traditional RGB cameras. However, reconstructing high-quality images from the spike input under low-light conditions remains challenging. Conventional learning-based methods often rely on the synthetic dataset as the supervision for training. Still, these approaches falter when dealing with noisy spikes fired under the low-light environment, leading to further performance degradation in the real-world dataset. This phenomenon is primarily due to inadequate noise modelling and the domain gap between synthetic and real datasets, resulting in recovered images with unclear textures, excessive noise, and diminished brightness.
To address these challenges, we introduce a novel spike-to-image reconstruction framework SpikeCLIP that goes beyond traditional training paradigms. Leveraging the CLIP model's powerful capability to align text and images, we incorporate the textual description of the captured scene and unpaired high-quality datasets as the supervision. 
Our experiments on real-world low-light datasets U-CALTECH and U-CIFAR demonstrate that SpikeCLIP significantly enhances texture details and the luminance balance of recovered images. Furthermore, the reconstructed images are well-aligned with the broader visual features needed for downstream tasks, ensuring more robust and versatile performance in challenging environments.
\end{abstract}

\begin{links}
    \link{Code}{https://github.com/chenkang455/SpikeCLIP}
\end{links}

\nnfootnote{\textsuperscript{$*$} Corresponding authors.}

\section{Introduction}
Spike cameras \cite{spikecamera}, drawing inspiration from the retina, represent an innovative method for capturing visual information by continuously converting light intensity into the binary spike stream. Unlike traditional cameras, which have a fixed exposure period and low capture frequency, spike cameras operate at an ultra-high speed with each pixel continuously monitoring the incoming light. Once the accumulated light reaches a certain threshold, a spike is emitted. This sampling mechanism enables the spike camera to record visual features at a remarkable rate of 40kHz, making it exceptionally suited for the high-speed imaging task. Additionally, spike cameras offer advantages such as lower bandwidth usage and higher dynamic range compared to traditional high-speed cameras like the Phantom. Nevertheless, the critical challenge of spike-to-image conversion remains to be solved since the binary quantized neuromorphic bitstream produced by the spike camera is not perceivable by human beings.

\begin{figure*}
    \centering
    \includegraphics[width=1\linewidth]{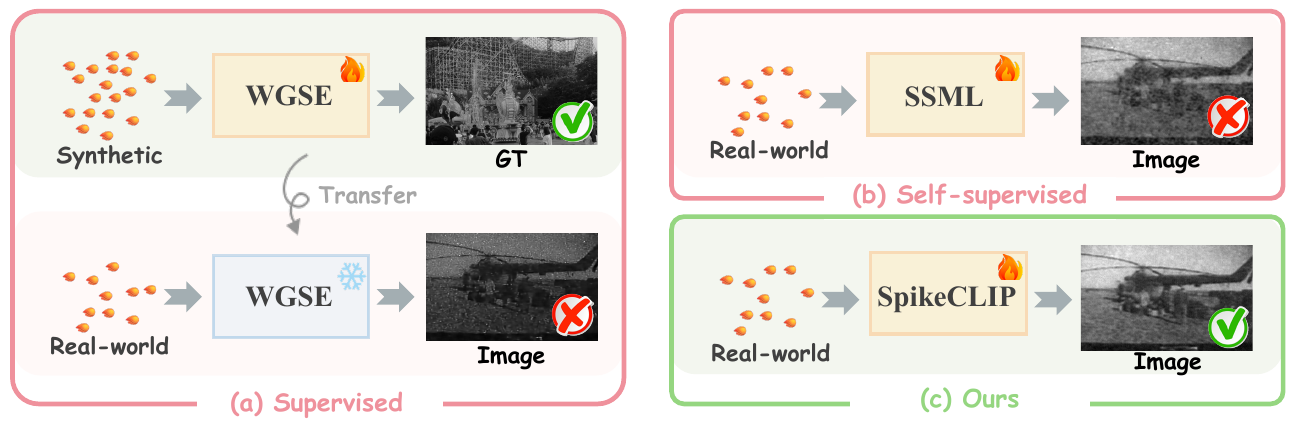}
    \caption{Illustration of the advantages of our method. While previous learning-based approaches struggle with real-world data under extreme conditions, such as low-light scenarios, our proposed SpikeCLIP successfully reconstructs high-quality images.}
    \label{fig:top_figure}
\end{figure*}

In recent years, various methods have been developed to map spike inputs to high-quality images. Early model-based approaches, designed based on the spike sampling principle \cite{tfp_tfi}, synaptic plasticity \cite{STDP_zheng}, and retina model \cite{zhu2020retina}, struggled with generalization and consistent reconstruction quality due to numerous adjustable parameters. To address these issues, learning-based methods like WGSE \cite{wgse} trained the convolutional network on the synthetic dataset with spike-image pairs, but the performance of these approaches degrades under the real-world low-light environment due to the dataset domain gap problem as illustrated in \cref{fig:top_figure}(a). Self-supervised methods have been explored as a solution for this challenge, but existing frameworks like SSML \cite{bsn_chen} and SJRE \cite{red_chen} heavily rely on pseudo-labels from TFI \cite{tfp_tfi}, which significantly degrade in quality under the low-light scene, leading to poor reconstruction results, as shown in \cref{fig:top_figure}(b).

To sum up, existing methods for spike-based image reconstruction fail to accurately recover sharp texture features under the low-light scene, highlighting the need for a novel paradigm that goes beyond previous approaches. This leads to a critical question: \textit{\textbf{what other signals can be used to supervise the spike-based image reconstruction network training besides the sharp image?}}

To this end, we propose the SpikeCLIP that leverages the text description as the supervision signal for the spike-based image reconstruction task.  We utilize the Contrastive Language-Image Pre-training (CLIP)~\cite{radford2021learning} model to extract the textual category information of the captured object and the feature representation of well-recovered images from our constructed high-quality image dataset. Benefiting from these designs, our proposed SpikeCLIP can reconstruct texture-rich, noise-free, and brightness-balanced images as shown in \cref{fig:top_figure}(c). Our proposed framework consists of three stages: (1) \textbf{\textit{Coarse Reconstruction}}, we utilize a lightweight network to map the spike input to an initial reconstruction output; (2) \textbf{\textit{Prompt Learning}}, learnable prompts are optimized to capture the distribution of high-quality and low-quality recovered images; (3) \textbf{\textit{Refinement}}, we leverage class-label features and learned prompts of high-quality images to enhance the network restoration performance further. 
We conduct quantitative and qualitative analyses on the U-CALTECH and U-CIFAR datasets, with experiments showing that our proposed SpikeCLIP achieves remarkable restoration performance under low-light conditions. The main contributions of this paper are as follows:
\begin{itemize}
    \item We introduce a novel spike-based image reconstruction framework tailored to the low-light environment, which leverages the CLIP model to supervise the network training by the class label of the captured object and the features of high-quality images.
    \item We design a high-quality image generation pipeline and demonstrate that a lightweight reconstruction network is sufficient for the spike-to-image task when the supervision signal is weak.
    \item Our method surpasses previous methods in reconstruction quality on the U-CALTECH and U-CIFAR datasets.
\end{itemize}

\begin{figure*}
    \centering
    \includegraphics[width=1.0\linewidth]{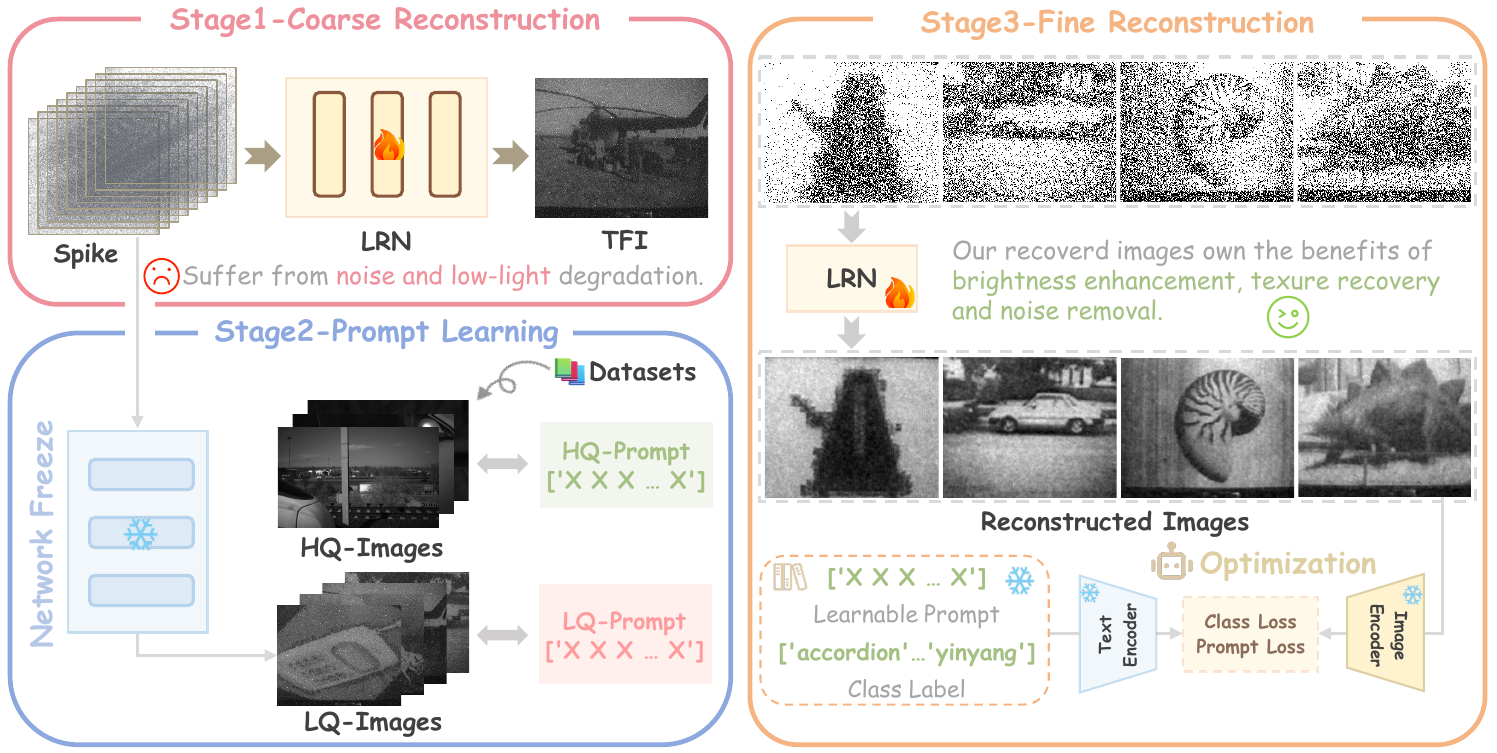}
    \caption{The overall framework of our three-stage spike-based image reconstruction method.}
    \label{fig:pipeline}
\end{figure*}

\section{Related Works}
\subsection{Spike Camera Image Reconstruction}
Spike-based image reconstruction methods can be roughly categorized into three types: model-based methods, supervised methods, and self-supervised methods. 
\subsubsection{Model-based methods.}
Model-based methods aim to build the relationship between the spike stream input and the sharp image output based on the spike camera sampling mechanism and biological principles.  \citet{tfp_tfi} proposed Texture from Play Back (TFP) and Texture from ISI (TFI) two methods to reconstruct the sharp frame from spike streams based on virtual exposure imaging principles and inter-spike interval analysis.  \citet{STDP_zheng} introduced a spike reconstruction framework based on the principles of short-term synaptic plasticity by exploring the temporal regularity of the spike stream. 

\subsubsection{Supervised methods.} Supervised learning methods are designed to train networks on paired synthetic datasets, which are subsequently used for inference in real-world scenarios.
  \citet{she2022spikeformer} introduced SpikeFormer, a hierarchical architecture encoder that progressively exploits multi-scale temporal and spatial features to reconstruct dynamic scenes from binary spike streams.  \citet{zhu2023recurrent} proposed a Recurrent Spike-based Image Restoration framework designed to reconstruct sharp frames from spike inputs under general illumination conditions.  \citet{zhao2024boosting} boost image reconstruction from the perspective of the spike fluctuations and design a spike representation module for fluctuation suppression.

\subsubsection{Self-supervised methods.} Self-supervised methods are proposed to overcome the domain-gap problem in supervised learning, enabling direct training in real-world scenarios without ground truth.  \citet{bsn_chen} constructed a self-supervised framework based on distillation learning, where the student model learns from the pseudo-labels provided by the blind-spot self-supervised denoising network (BSN).  \citet{red_chen} proposed a joint optimization framework for the BSN-based spike reconstruction network and the optical flow network, thus providing the mutual constraint for training in the real world.

\subsection{CLIP in Low-level Vision}
CLIP~\cite{radford2021learning} has been widely applied in various downstream recognition tasks owing to its robust capability to build the relationship between textual and visual features. However, there remains significant potential for exploring its application in low-level vision.  \citet{yang2024ldp} leverage the CLIP model to estimate accurate blur maps from dual-pixel pairs in an unsupervised manner to aid in enhancing the quality of deblurred images.  \citet{clip-list} introduces a novel unsupervised low-light image enhancement framework CLIP-LIT, which utilizes a prompt learning framework to enhance backlit images by imposing the network to learn the bright image features represented by the learnable prompt.  \citet{clip-denoise} proposed an asymmetrical encoder-decoder denoising network, focusing on transfer CLIP for generalizable image denoising.
 \citet{event-clip} designed a CLIP-based framework that leverages class prompt features to perform event reconstruction and classification simultaneously without the requirement of the labels.

\section{Preliminaries}
\subsection{Spike Camera Principle}
In this subsection, we detail the working principle of the spike camera. Initially, the camera photon receptor captures incoming photons and converts the light intensity into a voltage signal, which is continuously accumulated by an integrator. Whenever the voltage of the integrator exceeds the predefined threshold $\Theta$, the signal is reset and the spike camera emits a spike. Mathematically, the spike generation process can be formulated as follows:
\begin{equation}
\int_{0}^{t} I(s) ds \geq \Theta, \label{equ:spike_camera}
\end{equation}
where $t$ represents the spike firing time and $I(s)$ denotes the latent sharp frame at time $s$. 
The spike camera employs an asynchronous triggering mechanism with synchronous readout. Therefore, capturing a scene for $T$, we can obtain a spike stream with $K$ frames, denoted as $\mathcal{S} \in \{0,1\}^{K\times H \times W}$ with $H$ and $W$ represent the height and width respectively.

\begin{figure*}
    \centering
    \includegraphics[width=0.9\linewidth]{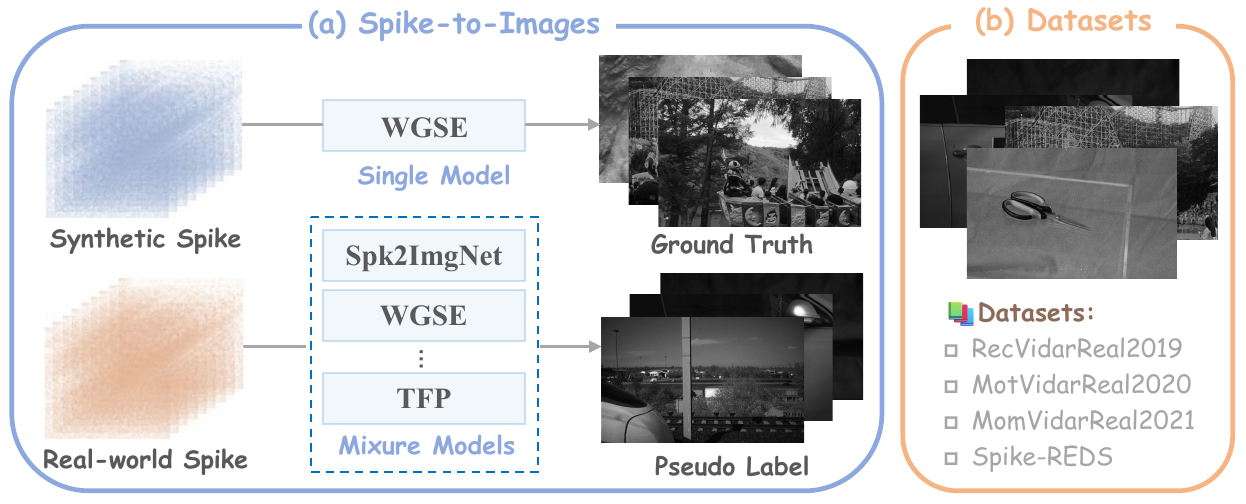}
    \caption{The framework of our designed HQ images generation pipeline.}
    \label{fig:hq_process}
\end{figure*}

\subsection{Motivation Clarification}
Given the spike stream $\mathcal{S}$, our objective  is to train the spike-based image \textbf{Recon}struction \textbf{Net}work \textbf{(Recon-Net)} under the low-light scenario, \ie, establishing the mapping from the sparse spike input to the sharp image $I$:
\begin{equation}
    I = \text{Recon-Net}(\mathcal{S};\theta),
\end{equation}
where $\theta$ represents the network parameters to be learned.

To optimize the parameters of the Recon-Net, previous supervised methods  \cite{spk2img,wgse,zhao2024boosting} utilize spike-sharp pairs for training, which can be mathematically represented as follows:
\begin{equation}
    \theta^* = \arg \min_{\theta} \mathcal{L}(\text{Recon-Net}(\mathcal{S};\theta), I_{\text{gt}}), \label{equ:train_recon}
\end{equation}
with $\mathcal{L}$ denoting the loss function designed to measure the difference between the reconstructed image and the ground truth sharp image $I_{\text{gt}}$, and $\theta^*$ represents the optimized network parameters. While SSML  \cite{bsn_chen} is the self-supervised method, its framework remains largely the same, except that it substitutes the ground-truth image $I_{\text{gt}}$ with the TFI reconstruction pseudo-label and the Recon-Net is the self-supervised denoising network BSN.

In this paper, we propose a new learning-based framework where the supervision signal is not the ground-truth sharp image but rather the class label and unpaired high-quality datasets encoded by the CLIP text encoder, which can be formulated as follows:
\begin{equation}
    \theta^* = \arg \min_{\theta} \mathcal{L}(\text{Recon-Net}(\mathcal{S};\theta), cls,D),
\end{equation}
where $cls$ represents the class label of the spike stream $\mathcal{S}$ and $D$ represents the unpaired high-quality datasets.

\section{Methods}
\subsection{Framework Overview}
Our framework consists of coarse reconstruction, prompt learning, and fine reconstruction three stages as shown in \cref{fig:pipeline}. In the coarse reconstruction, our designed network is trained to map the spike input to the TFI result. In prompt learning, we design a high-quality image generation pipeline and utilize learnable prompts to capture the distributions of high-quality (HQ) and low-quality (LQ) images. In the fine reconstruction, the network is fine-tuned to enhance the final output, guided by class-label features and the prompts learned from HQ images.

\subsection{Coarse Reconstruction} 
TFP and TFI \cite{tfp_tfi} are both effective and straightforward spike reconstruction algorithms, which can serve as the supervision signal for coarse reconstruction as in \cref{equ:train_recon}. However, TFP relies on a virtual exposure window, which requires a relatively long window in low-light scenes to capture sufficient spike information, resulting in severe motion blur. Conversely, TFI is not constrained by this window, making it suitable for low-light image reconstruction.

Given the spike stream $\mathcal{S}$ over the duration $T$, TFI initially calculates the Inter-Spike Interval (ISI) by searching for the spike firing time $t^{-}$ and $t^{+}$ before and after the middle time of each pixel, formulated as:
\begin{equation}
\text{ISI}_t = 
\begin{cases} 
    t^{+} - t^{-}, & \text{if } \exists\{t^{+}, t^{-}\} \in [0, T], \\
    \infty, & \text{otherwise}.
\end{cases}
\end{equation}
After calculating the ISI, the TFI image reconstruction result can be obtained based on the spike sampling principle depicted in \cref{equ:spike_camera}, \ie, $I_{\text{tfi}} = \Theta / \text{ISI}$. 

Regarding the architecture of the Recon-Net, since our method does not have ground truth as a strong constraint, we design a \textbf{L}ightweight \textbf{R}econstruction \textbf{N}etwork (LRN) composed of several convolutional layers as in \cite{chen2024spikereveal} to accomplish this task, which is efficient and sufficient as demonstrated by our further experiments. While the spike stream remains sparse under low-light conditions, we employ the voxel technique widely used in previous event-based vision tasks  \cite{E-CIR,trmd} to squeeze the input length of the spike sequence.

To sum up, the coarse reconstruction stage serves two main purposes. First, it enables the network to learn the fundamental mapping from the spike stream to a preliminary image representation. Second, the initial reconstruction result helps to minimize the sample distance between HQ and LQ images, aiding the learnable prompts to capture more distinctive features.

\subsection{Prompt Learning} 
We first construct a processing pipeline to generate HQ images, which are rich in texture features and balanced in brightness. Subsequently, we train the learnable prompts to capture and distinguish the features of HQ and LQ images.

\subsubsection{HQ Images Generation Pipeline.}
While we aim to train the LRN based on unpaired HQ datasets, the process of constructing such a dataset remains unsolved. Although synthetic datasets like Spike-REDS \cite{spk2img} contain sufficiently sharp images, there is a disparity in the brightness distribution pattern between these synthetic images and real-world images recovered by the spike camera. To overcome it, we construct the HQ dataset from synthetic and real two parts with the pipeline illustrated in \cref{fig:hq_process}.

For the synthetic dataset Spike-REDS, we reconstruct HQ images based on the best-performing method WGSE \cite{wgse} on this dataset. For real-world datasets captured under normal light, such as RecVidarReal2019 \cite{zhu2020retina}, MotVidarReal2020 \cite{motion_estimation_zheng}, and MomVidarReal2021 \cite{zheng2023spikecv}, we design a mixture model that combines algorithms like TFP, TFI, Spk2ImgNet, SSML and WGSE to generate high-quality images together. Among them, image $I_{hq}$ with the best non-reference metric NIQE \cite{niqe} score is added to the HQ-Images dataset, \ie:
\begin{equation}
I_{hq} = \arg \min_{m \in \mathcal{M}} \text{NIQE}(I_m)
\end{equation}
where $\mathcal{M}$ represents the set of methods and $I_m$ denotes the image reconstructed by method $m$.

\subsubsection{Prompt Optimization.} 
Once the high-quality dataset is constructed, we use the coarse reconstruction images by the LRN under the low light as the low-quality dataset, forming a pair of positive and negative sample datasets. Subsequently, similar to CLIP-LIT  \cite{clip-list}, we employ the learnable prompt framework based on CoOp \cite{coop} to obtain the learnable prompt representations of the positive and negative datasets.

Given the high-quality image $I_{hq}$ with its corresponding learnable prompt $T_{hq}$, and the low-quality image $I_{lq}$ reconstructed by the LRN with its learnable prompt $T_{lq}$. We utilize the CLIP image encoder $\Phi_{\text{image}}$ to process each image $I$ in the HQ and LQ datasets, obtaining the encoded image features aligned with the CLIP text encoder $\Phi_{\text{text}}$ output. Next, we formulate a binary classification problem to obtain the effective representation of the learnable prompts, \ie, the high-quality image should be aligned with the high-quality prompt and vice versa. We feed the learnable prompts into the text encoder and optimize them based on the cross-entropy loss function, formulated as follows:

\begin{align}
\mathcal{L}_{\text{initial}} &= \text{CrossEntropy}(y, \hat{y}), \\
\hat{y} &= \frac{e^{\Phi_{\text{image}}(I) \cdot \Phi_{\text{text}}(T_{hq})}}{ \sum_{i \in \{hq, lq\}} e^{\Phi_{\text{image}}(I) \cdot \Phi_{\text{text}}(T_{i})} },
\end{align}
where $I \in \{ I_{hq}, I_{lq} \}$ and $y$ is the label of the current image, 0 for the low-quality sample $I_{lq}$ and 1 for the high-quality sample $I_{hq}$.

\subsection{Fine Reconstruction} 
Previous research  \cite{event-clip} utilized the prompt `Image of a [cls]' to optimize the event reconstruction network. While the CLIP aligns text and images at a high-level semantic layer, directly applying it to the spike reconstruction low-level task introduces significant noise. To overcome this challenge, we utilize the learnable high-quality prompts along with class labels `X X...X [cls]' as the supervision.

Nevertheless, employing the coupled prompts for network optimization directly poses two main problems:
\begin{itemize}
    \item The class label is not coupled with the HQ prompt during the CLIP model optimization process, which can lead to misalignment when the coupled text is fed into the text encoder.
    \item The contrastive loss function designed for CLIP training cannot be applied to learnable prompts with two samples.
\end{itemize}

\begin{figure}
    \centering
    \includegraphics[width=0.85\linewidth]{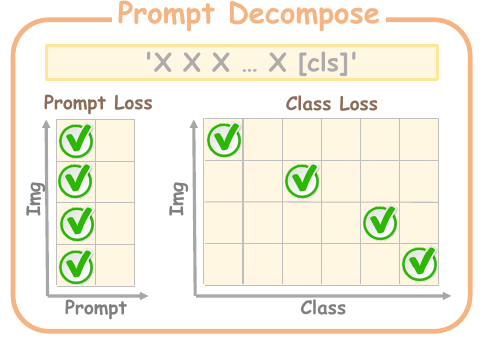}
    \caption{Prompt and class loss illustration.}
    \label{fig:loss}
\end{figure}

To the end, we decouple the prompts and design the prompt loss and the class loss to optimize the LRN reconstruction network separately, as illustrated in \cref{fig:loss}.

\subsubsection{Prompt Loss.}
With the learnable HQ and LQ prompts learned from the initialization stage, we can further optimize the LRN network like the previous study  \cite{clip-list}.  Specifically, we design the prompt loss to measure the alignment between the reconstructed images and the corresponding HQ prompts in the image-text space. The prompt loss $\mathcal{L}_{\text{prompt}}$ is formulated as:
\begin{equation}
\mathcal{L}_{\text{prompt}} = -\frac{e^{\Phi_{\text{image}}(I) \cdot \Phi_{\text{text}}(T_{hq})}}{\sum_{i \in \{hq, lq\}} e^{\Phi_{\text{image}}(I) \cdot \Phi_{\text{text}}(T_{i})}},
\end{equation}
where $I$ represents the reconstructed image.

\subsubsection{Class Loss.}
The input spike stream contains limited information under the low light condition. To this end, we utilize the class label `[cls]' of the captured object to supervise the training of the LRN, which aims to guide the network in learning high-level semantic features of the captured scenes from sparse inputs. Given that CLIP utilizes the InfoNCE loss  \cite{infonce} for aligning image-text pairs during training, we adopt a similar strategy by employing InfoNCE to optimize the LRN. Given a batch of spike stream $\mathcal{S} \in \{0,1\}^{B\times K \times H \times W}$, the loss function is formulated as follows:
\begin{equation}
\mathcal{L}_{\text{class}} = -\sum_{i=1}^{B} \log \frac{e^{(\Phi_{\text{image}}(I_i) \cdot \Phi_{\text{text}}(T_{c_i})) / \tau}}{\sum_{j=1}^{B} e^{(\Phi_{\text{image}}(I_i) \cdot \Phi_{\text{text}}(T_{c_j})) / \tau}},
\end{equation}
where $I_i$ represents the reconstructed image from the $i$-th spike input, $T_{c_i}$ is the text feature corresponding to the predicted class $c_i$, and $\tau$ is a temperature parameter that controls the sharpness of the contrast.

\subsubsection{Total Loss.}
The total loss $\mathcal{L}_{\text{total}}$ combines the prompt loss $\mathcal{L}_{\text{prompt}}$ and the class loss $\mathcal{L}_{\text{class}}$ as a weighted sum, formulated as:

\begin{equation}
\mathcal{L}_{\text{total}} = \mathcal{L}_{\text{class}} + \lambda \mathcal{L}_{\text{prompt}},
\end{equation}
where $\lambda$ is the hyper-parameter controlling the contribution of the prompt loss set to 100 in this task.

\begin{figure*}[t]
    \centering
    \includegraphics[width=1\linewidth]{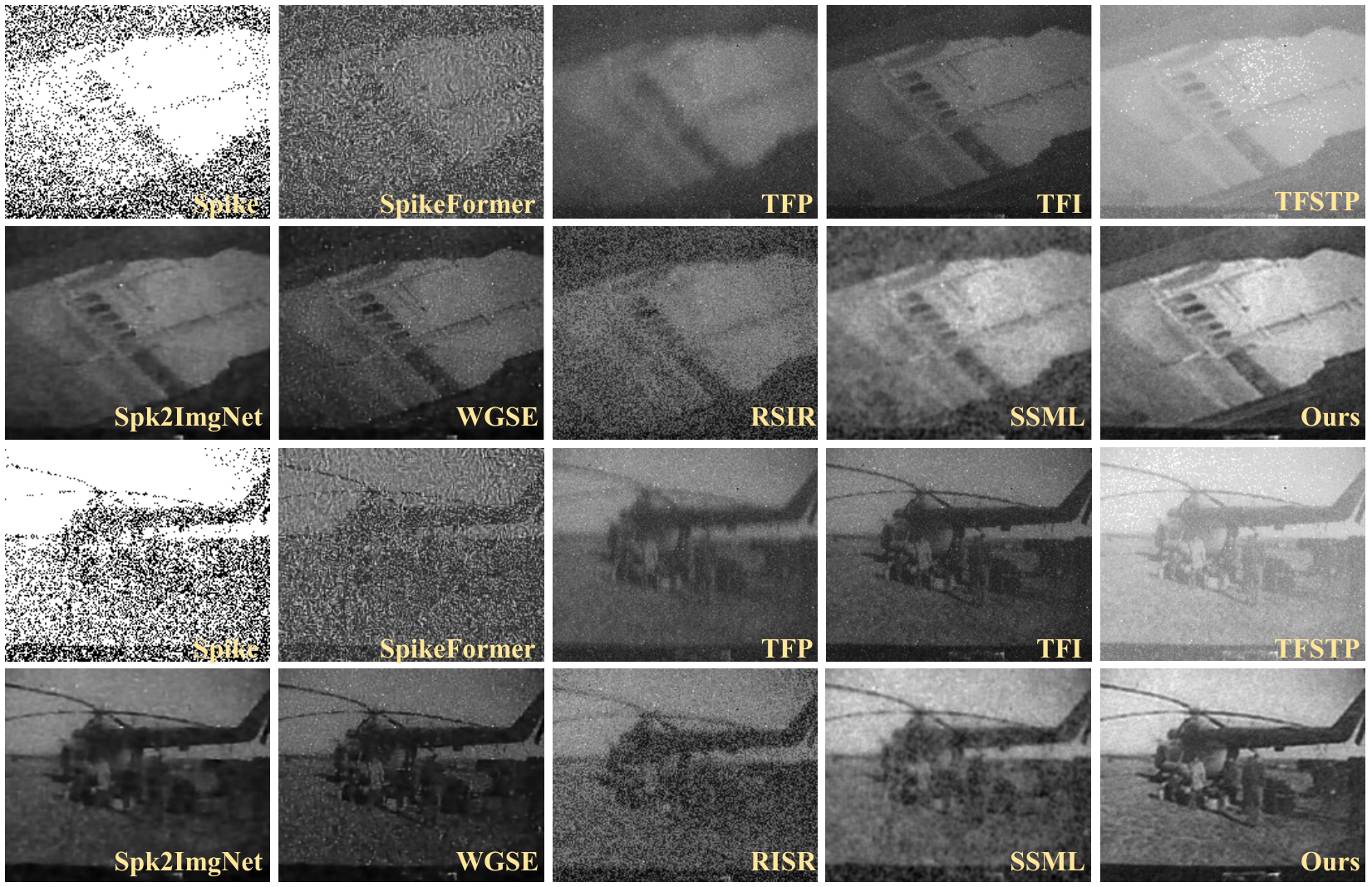}
    \caption{Visual comparison of our method with previous methods on the U-CALTEHC dataset.}
    \label{fig:exp}
\end{figure*}

\section{Experiments}
\subsection{Datasets}
We conduct both quantitative and qualitative experiments on the UHSR dataset  \cite{zhao2024recognizing}, which includes real-world spikes captured by the spike camera under low light from ultra-high-speed moving objects of the CALTECH  \cite{caltech101} and CIFAR  \cite{cifar} datasets, referred to as U-CALTECH and U-CIFAR respectively. Each dataset contains 5,000 spike-label training pairs and 1,000  test pairs, with $250 \times 250$ spatial resolution. We train the LRN network on the training sets and evaluate the image-text quality and classification performance on the test sets.


\subsection{Training Details}
We conduct the comparative experiment based on the Spike-Zoo \cite{spikezoo} repository, while the overall experimental framework is built on the PyTorch platform and trained based on an NVIDIA 4090 GPU. We employ the Adam optimizer with a learning rate of 4e-5 on the U-CALTECH dataset and 1e-4 on the U-CIFAR dataset. The training process consists of 25 total epochs: 5 epochs for coarse reconstruction, 1 epoch for prompt optimization, and 19 epochs for fine reconstruction. We employ the voxelization technique to convert the spike stream from an initial length of 200 to 50.

Since paired spike-sharp datasets are not available in real-world scenarios, we use non-reference metrics, NIQE  \cite{niqe}, BRISQUE  \cite{brisque}, and PIQE  \cite{piqe}, to evaluate the image reconstruction quality. In addition, we conduct a more comprehensive comparison by measuring the model parameter size (M), FLOPs (G), and time delay (ms) of each method. The FLOPs was calculated based on the standard spike stream with an input size of $250\times 400$ and the time delay was estimated by averaging 100 runs for each method on an NVIDIA 4090 GPU.

\begin{table*}[t]
\centering
\resizebox{\textwidth}{!}{
\begin{tabular}{ccccccccccccc} 
\toprule[2pt]
\multirow{2}{*}{Methods } & \multicolumn{3}{c}{U-CALTECH }  &  & \multicolumn{3}{c}{U-CIFAR} &  & \multirow{2}{*}{Params (M)} & \multirow{2}{*}{FLOPs (G)} & \multirow{2}{*}{Latency (ms)} \\ 
\cmidrule{2-4}\cmidrule{6-8}
  & NIQE $\downarrow$ & BRISQUE $\downarrow$ &  PIQE $\downarrow$  &  & NIQE $\downarrow$ & BRISQUE $\downarrow$ & PIQE $\downarrow$ &  & & & \\ 
\midrule
TFP    & 13.276  & 27.268   & 11.201   &  & 12.889  & 32.751  & 13.464  &  & /  & / & / \\
TFI   & 11.846  & 25.198   & \underline{4.762}    &  & 12.166  & 36.661  & \underline{7.133}   &  & /  & / & / \\
TFSTP  & 11.221 & \underline{23.446} & 12.460  &  & 11.872  & \underline{27.215}  & 13.556  &  & /  & / & / \\ 
\midrule
SSML  & 9.887  & 42.916   & 27.704   &  & 9.474   & 44.576  & 29.659  &  & 1.192  & 386 & 184 \\
SpikeFormer & 14.750  & 72.366   & 30.880   &  & 31.681  & 81.694  & 35.944  &  & 7.581  & 67.8 & 28.7 \\
RSIR & 28.302  & 111.673  & 15.047   &  & 45.640  & 122.826 & 18.013  &  & \underline{0.255}  & \textbf{6.461}  & \underline{7.18} \\
Spk2ImgNet & 11.120  & 74.275   & 76.348   &  & 11.114  & 81.411  & 92.902  &  & 3.904  & 1000 & 70.1 \\
WGSE  & \underline{7.449}  & 24.683   & 21.053   &  & \underline{9.458}   & 48.438  & 36.027  &  & 3.806  & 415 & 40.8 \\ 
\midrule
\rowcolor{gray!20}
\textbf{Ours}  & \textbf{4.657}  & \textbf{14.955}   & \textbf{3.265}   &  & \textbf{5.383}  & \textbf{25.831}  & \textbf{4.298}  &  & \textbf{0.186}  & \underline{18.562} & \textbf{0.540} \\
\bottomrule[2pt]
\end{tabular}
}
\caption{Comparison of our method with SOTA methods on  U-CALTECH and U-CIFAR datasets. We use the \textbf{bold} and \underline{underline} to distinguish the best and second-best results.}
\label{tab:exp}
\end{table*}

\subsection{Experimental Results}
We compare our method with previous state-of-the-art (SOTA) spike-based image reconstruction methods, including model-based approaches TFP  \cite{tfp_tfi}, TFI  \cite{tfp_tfi}, and TFSTP  \cite{STDP_zheng}, supervised learning methods designed for normal lighting conditions including Spk2ImgNet  \cite{spk2img}, SpikeFormer  \cite{she2022spikeformer}, and WGSE \cite{wgse} as well as methods tailored for low-light scenarios RSIR  \cite{zhu2023recurrent}, and the self-supervised learning approach SSML  \cite{bsn_chen}. 

We conduct qualitative experiments on the U-CALTECH and U-CIFAR datasets as detailed in \cref{tab:exp}, which demonstrates that our method outperforms SOTA on both datasets, delivering superior performance with lower parameters, computational complexity, and latency. Specifically, compared to the second-best method for each metric, our method reduces the NIQE score by 37.9\% (compared to WGSE), the BRISQUE score by 36.2\% (compared to TFSTP), and the PIQE score by 31.4\% (compared to TFI) and by 43.1\%, 5.1\%, and 39.7\% respectively on the U-CIFAR dataset.

We further conduct a visual comparison of our method with SOTA on the U-CALTECH dataset, as shown in \cref{fig:exp}. In particular, our method demonstrates superior noise reduction and detail preservation across recovered images. Unlike previous methods producing results with significant noise and blurry details, our approach consistently yields sharper and more accurate reconstructions with brightness balance, which complements the quantitative improvements as discussed earlier.

\subsection{Ablation Study}
In the ablation study, we quantitatively analyze the impact of each component and the model structure from the image quality and classification accuracy two aspects. Since we have aligned the output of the LRN with the text description through the $\mathcal{L}_{\text{class}}$ loss during training, we can leverage the CLIP model to directly classify the class of the captured scene, which assists in evaluating the overall quality of the reconstructed images.

\subsubsection{Effectiveness of Each Component.}
We conducted ablation experiments to evaluate the effectiveness of the first-stage TFI coarse reconstruction, the class loss $\mathcal{L}_{\text{class}}$, and the prompt loss $\mathcal{L}_{\text{prompt}}$ as shown in \cref{tab:abl1}. 
While the TFI image feature is not aligned with the CLIP text space, its classification accuracy is relatively low and the image quality suffers due to significant noise and unbalanced luminance.
After introducing the classification loss function $\mathcal{L}_{\text{class}}$, we observe substantial improvements across all metrics, benefiting from the high-level features provided by the class features of the captured scene and the alignment with the CLIP text encoder. Additionally, we observe that the initial coarse reconstruction is beneficial for network performance, likely due to the improved initialization of network parameters.
Finally, the model achieves the best performance when all components are active, further demonstrating the benefits of incorporating learnable HQ prompts.

\subsubsection{Effectiveness of Network Architecture.} 
While we claim that our designed LRN is well-suited for spike-based image reconstruction under this setting, we further designed a spike-based reconstruction model based on UNet  \cite{UNet} and compare it with the LRN as shown in \cref{tab:abl2}. Comparison demonstrates the effectiveness of our LRN model, confirming that a lightweight network architecture is not only sufficient but also efficient while the supervision signal is weak.

\begin{table}
\centering
\resizebox{\linewidth}{!}{
\begin{tabular}{cccccc} 
\toprule[2pt]
TFI & $\mathcal{L}_{\text{class}}$  & $\mathcal{L}_{\text{prompt}}$  & NIQE $\downarrow$  & BRISQUE $\downarrow$ & ACC $\uparrow$    \\ 
\midrule
\ding{51} &     &     & 15.389 & 29.957   & 1.60   \\
    & \ding{51} &     & 6.043  & 15.821   & 58.90  \\
\ding{51} & \ding{51} &     & 4.799  & 18.748   & 63.00  \\
\ding{51} & \ding{51} & \ding{51} & \textbf{4.656}  & \textbf{14.955}   & \textbf{64.00}  \\
\bottomrule[2pt]
\end{tabular}
}
\caption{Ablation on the contribution of each component.}
\label{tab:abl1}
\end{table}

\begin{table}
\centering
\resizebox{\linewidth}{!}{
\begin{tabular}{ccccccc} 
\toprule[2pt]
Model & NIQE $\downarrow$& BRISQUE $\downarrow$& ACC $\uparrow$&  Params & Flops   \\ 
\midrule
UNet  &   25.94   &    69.64    & 61.70  &    19.2    &    46.46            \\
LRN   &  \textbf{4.657}    &   \textbf{14.95}     & \textbf{64.00}  &    \textbf{0.186}    &   \textbf{18.56}          \\
\bottomrule[2pt]
\end{tabular}
}
\caption{Ablation on the network architecture.}
\label{tab:abl2}
\end{table}

\section{Conclusion}
In summary, we propose a novel spike camera image reconstruction framework tailored for extreme conditions like low light. We design a CLIP-based three-stage coarse to refine reconstruction pipeline, leveraging label textual cues and unpaired high-quality datasets to guide network training. Experimental results on U-CALTECH and U-CIFAR datasets demonstrate that our method surpasses SOTA in recovering images with  details and balanced brightness.

\section*{Acknowledgments}
We sincerely appreciate Yuyan Chen (HUST) for her valuable suggestions and for polishing the figures. This work was supported by the National Natural Science Foundation of China (62422601, 62176003, 62088102, 62306015), the China Postdoctoral Science Foundation (2023T160015), the Young Elite Scientists Sponsorship Program by CAST (2023QNRC001), and the Beijing Nova Program (20230484362). 

\bibliography{aaai25}
\newpage

\end{document}